\documentclass[letterpaper]{article} 
\usepackage{aaai2027}  
\usepackage{microtype}
\usepackage{enumitem}
\usepackage{cuted}
\usepackage{capt-of} 
\usepackage{booktabs}
\usepackage{multirow}
\usepackage{booktabs}
\usepackage{amsmath}
\usepackage{cleveref}
\usepackage{adjustbox}
\usepackage{amssymb}
\usepackage{array}
\usepackage[table]{xcolor}   
\newcommand{\icon}[1]{\raisebox{-0.15em}{\includegraphics[height=1em]{#1}}}
\usepackage[hyphens]{url}  
\usepackage{graphicx} 
\usepackage{natbib}  
\usepackage{caption} 
\usepackage{algorithm}
\usepackage{algorithmic}

\usepackage{comment}

\usepackage{newfloat}
\usepackage{listings}
\DeclareCaptionStyle{ruled}{labelfont=normalfont,labelsep=colon,strut=off} 
\floatstyle{ruled}
\newfloat{listing}{tb}{lst}{}
\floatname{listing}{Listing}

\usepackage{booktabs}

\title{SAP-Nav: Spatial Semantic Representation Meets Active Perception\\for Hierarchical Open-Vocabulary Object Navigation}

\author{
    Xuetong Pei\textsuperscript{\rm 1, 2},
    Jian Liu\textsuperscript{\rm 2},
    Vidura Munasinghe\textsuperscript{\rm 2},
    Bo Miao\textsuperscript{\rm 3},
    U-Xuan Tan\textsuperscript{\rm 2},
    Wenrui Ding\textsuperscript{\rm 1},
    Na Zhao\textsuperscript{\rm 2}
}
\affiliations{
    \mbox{\textsuperscript{\rm 1}Beihang University \quad
    \textsuperscript{\rm 2}Singapore University of Technology and Design \quad
    \textsuperscript{\rm 3}AIML, Adelaide University}
}

\begin{document}
\maketitle
\begin{strip}
    \centering
    \includegraphics[width=\textwidth]{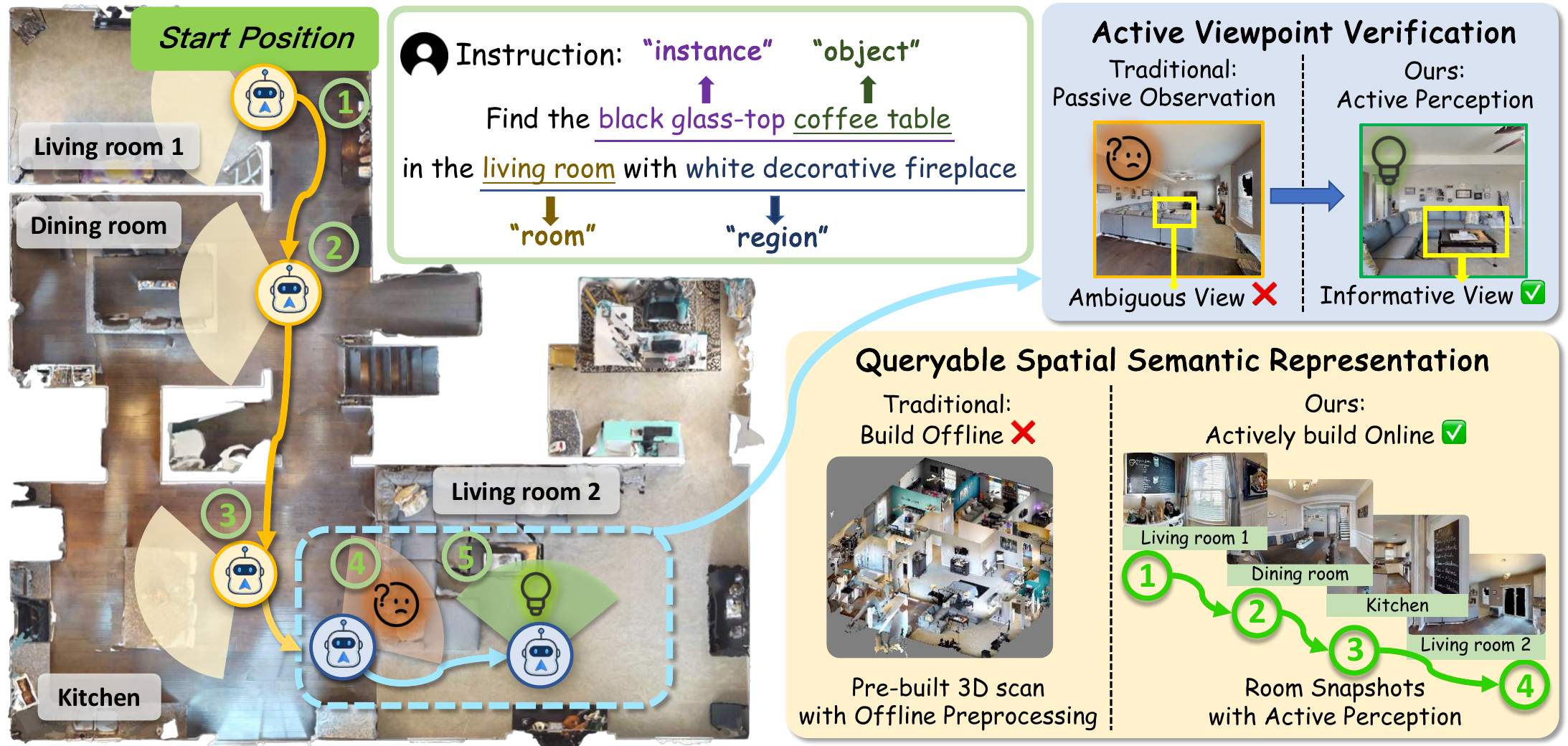}
    \captionof{figure}{
SAP-Nav introduces active perception into hierarchical OVON, where goals span scene-, room-, region-, and instance-level specifications.
Unlike passive methods relying on observations encountered during navigation, SAP-Nav actively seeks informative viewpoints to build a queryable spatial semantic representation online and identify the intended target.
}
\label{fig:teaser}
\end{strip}
\begin{abstract}

Hierarchical open-vocabulary object navigation (OVON) requires agents to follow free-form instructions that may specify targets through scene-, room-, region-, and instance-level cues in unseen environments. Although recent work LangMap has formalized this setting, reliably solving it under partial observations remains challenging: spatial grounding requires persistent environment-level evidence, whereas target verification requires clear and discriminative candidate views. We present SAP-Nav, a fully online, zero-shot framework that addresses both requirements through active perception. SAP-Nav incrementally constructs a Queryable Spatial-Semantic Representation from actively acquired room views, enabling spatial semantic queries from any explored location. It further employs Active Viewpoint Verification to assess whether the current observation provides sufficient evidence and, when necessary, reposition the agent to a more informative viewpoint before verifying candidates against category and attribute constraints. Although designed for hierarchical OVON, SAP-Nav supports both hierarchical and standard category-level OVON without task-specific training or precomputed scene maps. Experiments on LangMap and HM3D-OVON show that SAP-Nav achieves the overall best performance, including a 12.2\% improvement in SR over training-based methods on region-level navigation. Real-world robot experiments further demonstrate its practical feasibility. Code will be made publicly available upon acceptance. \pdfstartlink attr{/Border[0 0 0]} user{/Subtype/Link/A<</S/URI/URI(https://xuetongpei.github.io/SAP-Nav/)>>}\textcolor{blue}{Project Page}\pdfendlink.
\end{abstract}

\section{Introduction}
Open-vocabulary object navigation (OVON) requires embodied agents to navigate to language-specified targets in previously unseen environments using only onboard observations. Standard OVON benchmarks \cite{hm3dovon} and methods \cite{vlfm, uninavid} primarily formulate goals at the object-category level, where reaching any instance of the specified category constitutes success. Although this formulation enables standardized evaluation, it does not fully reflect how people naturally express navigation goals. For example, ``\textit{bring me the red bottle in my bedroom}'' specifies an object category together with room-level context and instance-level attributes. LangMap \cite{langmap} formalizes this broader setting by extending OVON evaluation to scene-level category goals and room-, region-, and instance-level constraints, as illustrated by the instruction example in Fig.~\ref{fig:teaser}. We refer to this setting as \emph{hierarchical OVON}, which spans broad category search to context-rich target disambiguation.

Reliably grounding goals across these levels requires complementary visual evidence under partial observations. Room context and specific room instances are best inferred from holistic observations that reveal the surrounding spatial layout and semantics, whereas instance-level constraints require clear, discriminative views of candidate objects. Passive exploration does not guarantee either type of evidence. Methods that rely on complete 3D scans \cite{hov-sg, spatialnav} assume information unavailable in fully online deployment, while methods that passively accumulate incidental onboard observations \cite{sg-nav, spatial-aware} may obtain incomplete or viewpoint-biased scene coverage. Similarly, a valid candidate may initially appear distant, occluded, or from an uninformative perspective, making direct verification unreliable. Hierarchical OVON therefore requires agents not only to retain structured spatial-semantic evidence, but also to actively acquire observations that are informative for both grounding and verification.

To address these challenges, we propose \textbf{SAP-Nav}, a fully online, zero-shot framework that applies active perception at two complementary stages. As illustrated in Fig.~\ref{fig:teaser}, SAP-Nav comprises a Queryable Spatial-Semantic Representation (QSSR) and Active Viewpoint Verification (AVV). QSSR is incrementally constructed during navigation through online room segmentation and actively acquired holistic room observations. These observations provide persistent evidence for assigning room-level semantics and disambiguating multiple rooms of the same type, \textit{e.g.}, the two living rooms shown in Fig.~\ref{fig:teaser}. The resulting representation supports spatial-semantic queries at any explored location. AVV complements this environment-level representation with candidate-level evidence acquisition. Rather than immediately verifying a detected candidate from the current observation, AVV first assesses view sufficiency and, when necessary, moves the agent to a more informative and feasible viewpoint before verifying the candidate against the category and attribute constraints in the instruction. Together, QSSR and AVV enable SAP-Nav to address hierarchical OVON without task-specific training or precomputed scene maps, while remaining applicable to standard category-level OVON. Our main contributions are:
\begin{itemize}
    \item We present SAP-Nav, a fully online, zero-shot framework designed for hierarchical OVON and applicable to standard category-level OVON.
    \item We introduce QSSR, which incrementally combines online room segmentation with actively acquired holistic room observations to assign room-level semantics, disambiguate rooms of the same type, and support persistent spatial-semantic queries at any explored location.
    \item We develop AVV, a closed-loop module that actively guides the agent to informative viewpoints when current observations are insufficient to verify candidates, enabling reliable identification of the intended target.
    \item We show through extensive experiments that SAP-Nav achieves the best overall performance across hierarchical and standard OVON settings, and validate its practical feasibility in real-world robot deployments.
\end{itemize}

    \label{sec:framework}
    \begin{figure*}[t]
        \centering
        \begin{adjustbox}{center}
            \includegraphics[width=\textwidth]{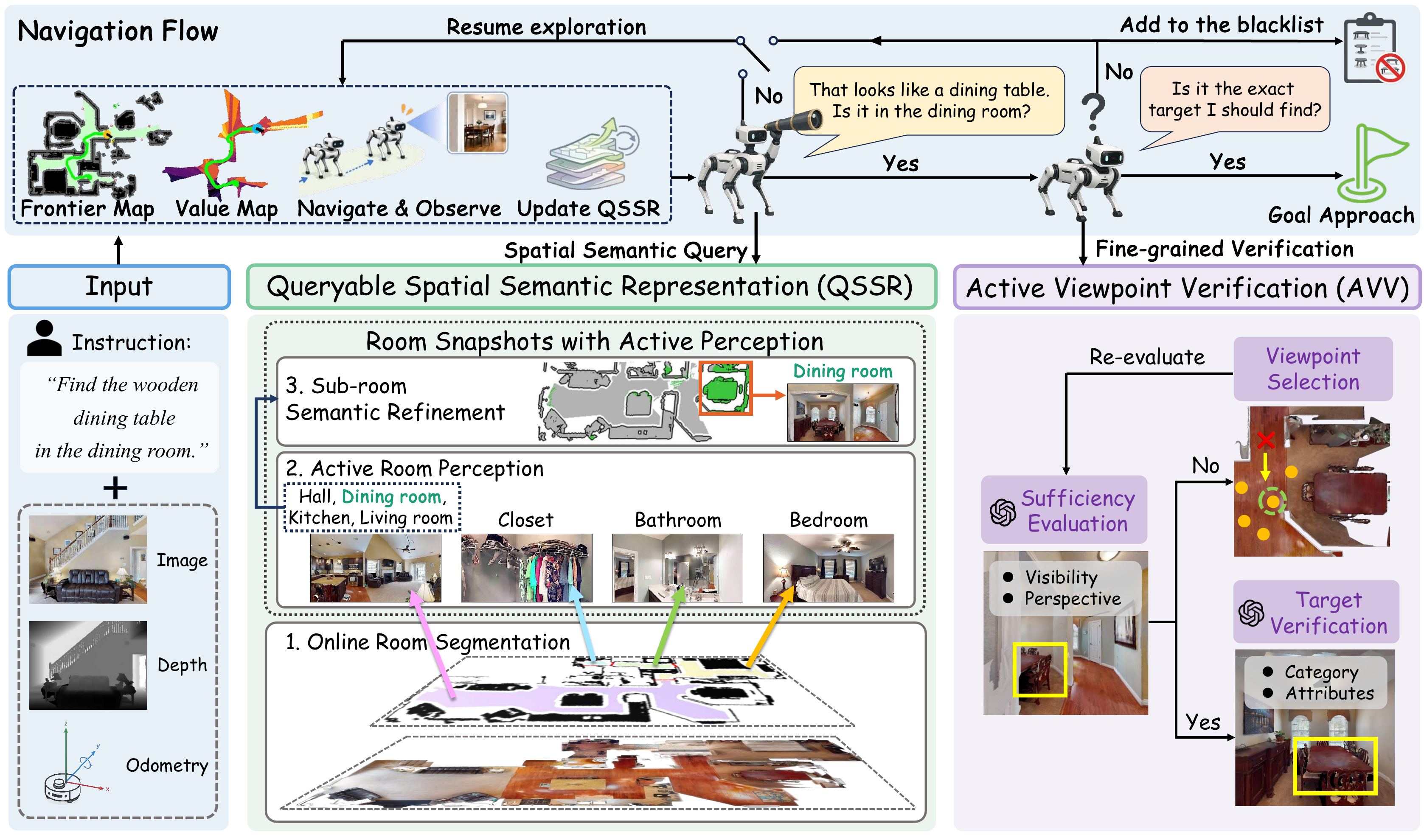
            }
        \end{adjustbox}
        \caption{
\textbf{The overview of our SAP-Nav}.
Given egocentric RGB-D observations, odometry, and an instruction at any granularity, SAP-Nav operates through two active-perception modules.
QSSR converts actively acquired online observations into a queryable spatial semantic representation, from room segmentation to sub-room semantics.
AVV closes the verification loop by assessing viewpoint sufficiency, actively repositioning to informative views, and verifying candidates against the goal category and attributes; rejected candidates are blacklisted and exploration continues.
}
        \label{fig:framework}
        \vspace{-0.2in}
    \end{figure*}
    
\section{Related Work}
\paragraph{Open-Vocabulary Object Navigation.}
OVON extends closed-set object navigation~\cite{hm3d,mp3d} to goals drawn from an open vocabulary~\cite{hm3dovon}, typically requiring the agent to find any matching instance of a category.
LangMap~\cite{langmap} broadens this setting to scene-, room-, region-, and instance-level instructions, better reflecting how users specify targets in practice.
Existing methods can be categorized by their reliance on task-specific navigation training.
Training-based approaches learn implicit scene and instruction representations and train navigation capabilities from large-scale navigation data \cite{hong2021vln, mousavian2019visual,ramrakhya2023pirlnav, wijmans2022ver,ye2021auxiliary,zengpoliformer, navfom}.
Training-free methods instead build on off-the-shelf VLMs for zero-shot navigation.
Early training-free methods focus on scene-level goals~\cite{vlfm,yu2023frontier}, while recent approaches accept richer instructions but still infer spatial constraints and instance attributes from passively acquired observations~\cite{longinstructnav,3dmem}.
These limitations motivate a fully online, zero-shot method that explicitly tackles multi-granularity instructions.

\paragraph{Semantic Scene Representations for Navigation.}
Semantic scene representations organize observations into spatial and semantic abstractions that support spatial reasoning and navigation.
Existing methods construct these representations either offline from complete scenes or online from onboard observations.
Offline methods~\cite{hov-sg, spatialnav} build hierarchical representations from complete 3D scans, providing rich global context but assuming prior access to the environment.
Online approaches primarily emphasize room geometry~\cite{hydra}, or encode room semantics implicitly through object-level features~\cite{spatial-aware} at the room level.
They therefore lack an explicit representation that associates room instances with semantic types during online exploration.
Constructing such a representation is challenging because room semantics is a holistic property whose evidence accumulates across multiple views. 
This motivates us to construct a spatial semantic representation online which encodes explicit room semantics.

\paragraph{Active Perception for Embodied tasks.}
Active perception refers to deliberately controlling sensing behavior, deciding what, how, when, and where to perceive, based on the current state of interpretation \cite{revisiting} and has recently gained renewed attention in related fields~\cite{activescene_recognition, activeVLN, activeOVMM, activevla}.
Object navigation is naturally suited to such deliberate viewpoint control, as each action of the agent changes the pose of the onboard camera, and thereby determines the available visual evidence.
Existing methods, however, typically select actions to make progress toward the goal, with visual observations acquired only as a byproduct of the resulting trajectory~\cite{openfrontier,ovsegdt}, leaving observation quality rarely treated as an explicit objective of action selection.
This motivates us to control viewpoints actively for perception in hierarchical OVON.

\section{Method}
\paragraph{Problem Formulation.}
We consider an agent navigating an unknown environment, which receives a natural-language goal $G$ at one of multiple granularities, from an object category to targets constrained by spatial context and discriminative attributes.
At each timestep $t$, it observes an egocentric RGB image $I_t$, depth $D_t$, estimates its pose $P_t$ from odometry, and executes an action $A_t$.
Using only egocentric observations, the agent must navigate to a target that satisfies $G$, and the task is considered successful when it stops within a predefined distance of that target.

\paragraph{Method Overview.}
As illustrated in Fig.~\ref{fig:framework}, SAP-Nav addresses hierarchical OVON through two stages that actively acquire visual evidence for spatial semantic understanding and target verification. 
Given the task inputs, SAP-Nav searches for the target using vision-language-guided frontier exploration~\cite{vlfm}. 
During exploration, the agent constructs Queryable Spatial-Semantic Representation (QSSR) online as a hybrid representation of a spatial semantic BEV map and actively acquired holistic room snapshots.
For each detected candidate, the agent queries QSSR to determine whether the candidate satisfies the room- and region-level constraints in $G$.
Only spatially consistent candidates trigger Active Viewpoint Verification (AVV), which actively improves the viewpoint when necessary and verifies the target against the category and attributes. 
Candidates rejected by AVV are blacklisted and exploration resumes, whereas the accepted one is approached.
\subsection{Queryable Spatial Semantic Representation}
QSSR is designed to support querying the room type at any explored space and using holistic room snapshots to distinguish among regions (\textit{i.e.}, multiple room instances of the same type).
It is a hybrid representation comprising a spatial semantic BEV map where each grid cell stores room type semantics, and a visual record that associates actively acquired room snapshots with corresponding room instances. 
It is built progressively through online room segmentation, active room perception, and sub-room semantic refinement, denoted as $S_1$, $S_2$ and $S_3$ respectively. 

\paragraph{Online Room Segmentation.}
QSSR first establishes a structure of room instances over the online obstacle map by partitioning the explored free space into rooms.
Doors provide natural boundaries for room segmentation, but viewpoints encountered during object navigation often yield sparse and biased door observations.
We therefore adapt the online room segmentation method~\cite{topology} to goal-driven exploration by accumulating door projections over a temporal window of $m$ frames.
Each door boundary indicates entry into a new room and provides the inward direction for capturing a holistic room snapshot. The resulting partition maps every explored BEV location to a room instance $R_k$, providing the structural basis for subsequent stages.

\paragraph{Active Room Perception}
Based on the geometric structure provided by $S_1$, the stage $S_2$ associates each room instance $R_k$ with a set of semantic types $C_k$ and a visual record $V_k$, denoted as $R_k \rightarrow (C_k,V_k)$. 
Since observations encountered during navigation may cover only a limited part of a room, $V_k$ is constructed from holistic room snapshots acquired at key exploration events, and $C_k$ is inferred from $V_k$. $S_2$ determines when to observe and how to orient the camera without introducing additional translational movement. Specifically, snapshots are acquired under three conditions:

(1) \emph{Initialization.}
The agent reuses the initial panorama captured at the start of each episode as $v_0 \in V_0$ for the starting room $R_0$.
(2) \emph{First entry.}
When the agent first enters a new room $R_k$, it faces the room interior along the inward normal of the entry door $d_k$ and rotates in place to capture $N_v$ frames, which are stitched into a room snapshot $v_k^0 \in V_k$.
(3) \emph{Expansion update.}
We capture a new stitched view to update the representation of a previously classified room $R_k$ whenever its area grows substantially relative to the area at the last classification.
The $n$-th new stitched view $v_{k_n} \in V_k$ is aimed at the centroid of the newly explored portion of $R_k$, so that the updated snapshot reflects the full extent of the room rather than only the area seen at first entry.

Each room snapshot $v_{k_n}$ is then fed into the VLM to predict a semantic type $c_{k_n}$, and the predictions from all snapshots of $R_k$ are accumulated in $C_k$:
 \begin{equation}
      c_k^n=\mathrm{VLM}_{\mathrm{cls}}(v_k^n),
      \qquad
      C_k=\bigcup_{v_k^n\in V_k}\{c_k^n\}.
  \end{equation}
Together, $C_k$ and $V_k$ form the room level semantic record of $R_k$. $C_k$ provides the semantic prior for sub-room refinement, while $V_k$ retains holistic visual evidence for distinguishing among different regions.

\paragraph{Sub-Room Semantic Refinement}
The stage $S_2$ assigns room type semantics to each segmented room, but this can be too coarse in open areas or under-segmented spaces where multiple functional areas are merged into one.
We therefore refine QSSR with a sub-room semantic probability field $\Phi$ over the BEV map to distribute $c_k \in C_k$ to sub-room areas within $R_k$.
For each image patch $p$, we use a pretrained scene encoder $\phi$~\cite{places} to produce a room type logit vector $\phi_t(p)$.
Since the label space of the scene encoder contains out-of-domain categories for indoor navigation, we impose a static prior that restricts predictions to an indoor room-type vocabulary $\mathcal{T}$, yielding
      $\tilde{\phi}_t(p)
      =
      \phi_t(p)\big|_{\mathcal{T}}
      \in \mathbb{R}^{|\mathcal{T}|}$.
Then we back-project $\tilde{\phi}_t(p)$ to its corresponding BEV grid, denoted as $u$, turning per observation of room probabilities into per grid room type.
For each cell $u$, evidence from patches observed across timesteps and viewpoints is accumulated as
\begin{equation}
    \label{eq:bayes-fusion}
   \Phi_t(u)=\sum_{i=1}^{t}\tilde{\phi}_i(u),
\end{equation}
which is updated throughout exploration and provides a normalized room type distribution at each $u$, thereby distinguishing functional areas in open or undersegmented spaces.
\paragraph{Spatial Semantic Query}
When a candidate object, denoted as $o$, is detected, the agent queries QSSR with BEV cells covered by its projected segmentation footprint, denoted as $\mathcal U(o)$, to check whether $o$ satisfies the spatial constraints in $G$. QSSR first determines the room type $\hat{c}(o)$ of $o$ as:
  \begin{equation}
  \hat{c}(o)=
  \begin{cases}
  c, & C_k=\{c\}, \\[3pt]
  \displaystyle
  \operatorname*{argmax}_{c\in C_k}
  \frac{1}{|\mathcal{U}(o)|}
  \sum_{u\in\mathcal{U}(o)}
  \Phi_t(u)_c,
  & |C_k|>1.
  \end{cases}
  \end{equation}
If $G$ refers to a region, the agent further uses $V_k$ as visual evidence for VQA to determine whether $R_k$ matches the region description in $G$.

\subsection{Active Viewpoint Verification}
After localizing the room or region specified by the instruction, the agent may encounter a spatially consistent candidate that is visible in the current observation, while the viewpoint provides insufficient evidence for reliable verification.
We therefore introduce AVV, comprising viewpoint sufficiency evaluation, viewpoint selection, and target verification.

\paragraph{Sufficiency Evaluation}
Given the current observation $I_t$, we assess whether the candidate can be reliably verified along two criteria:
(1) \emph{visibility}, assessing whether the candidate is clearly visible; and
(2) \emph{perspective}, assessing whether it is viewed from an angle that reveals sufficient identifying details~\cite{taven}.
We prompt the VLM with both the candidate crop and the full egocentric image $I_t$, which provide local appearance and surrounding context, respectively.
Each criterion is scored from 1 to 5, and their sum defines the sufficiency score $s \in \{2,\ldots,10\}$.
If $s < \tau$, the current view is considered insufficient, and the agent performs viewpoint selection to acquire a more informative observation; otherwise, it proceeds directly to target verification.

\begin{table*}[t]
\centering

\setlength{\tabcolsep}{4pt}
\renewcommand{\arraystretch}{1.25}
\resizebox{\textwidth}{!}{%
\begin{tabular}{l l cc|cc cc cc cc}
\toprule
\multirow{2}{*}{\textbf{Method}} & \multirow{2}{*}{\textbf{VLM}}
& \multicolumn{2}{c|}{\textbf{Single-Goal}}
& \multicolumn{2}{c}{\textbf{Scene}}
& \multicolumn{2}{c}{\textbf{Room}}
& \multicolumn{2}{c}{\textbf{Region}}
& \multicolumn{2}{c}{\textbf{Instance}} \\
\cmidrule(lr){3-4} \cmidrule(lr){5-6} \cmidrule(lr){7-8} \cmidrule(lr){9-10} \cmidrule(lr){11-12}
 & & SR\,$\uparrow$ & SPL\,$\uparrow$ & SR & SPL & SR & SPL & SR & SPL & SR & SPL \\
\midrule
\rowcolor{gray!15} \multicolumn{12}{l}{\textit{Training-based methods}} \\
PSL~{\scriptsize[ECCV'24]~\cite{psl}}          & -- & 6.6  & 1.8  & 6.0  & 1.4  & 6.6  & 1.9  & 7.3  & 2.1  & 6.4  & 1.9 \\
SenseAct-M~{\scriptsize[CVPR'24]~\cite{goat}}  & -- & 8.7  & 4.6  & 10.2 & 5.6  & 8.3  & 4.4  & 8.5  & 4.3  & 7.7  & 4.0 \\
MTU3D~{\scriptsize[ICCV'25]~\cite{mtu3d}}      & -- & 29.7 & 15.4 & 33.1 & 16.7 & 31.4 & 15.9 & 32.7 & 16.5 & 23.8 & 13.9 \\
Uni-NaVid~{\scriptsize[RSS'25]~\cite{uninavid}} & -- & 30.3 & 15.3 & 33.8 & 16.2 & 33.2 & 16.5 & 30.1 & 15.5 & 26.2 & 13.8 \\
PlaNaVid~{\scriptsize[arXiv'26]~\cite{langmap}} & \icon{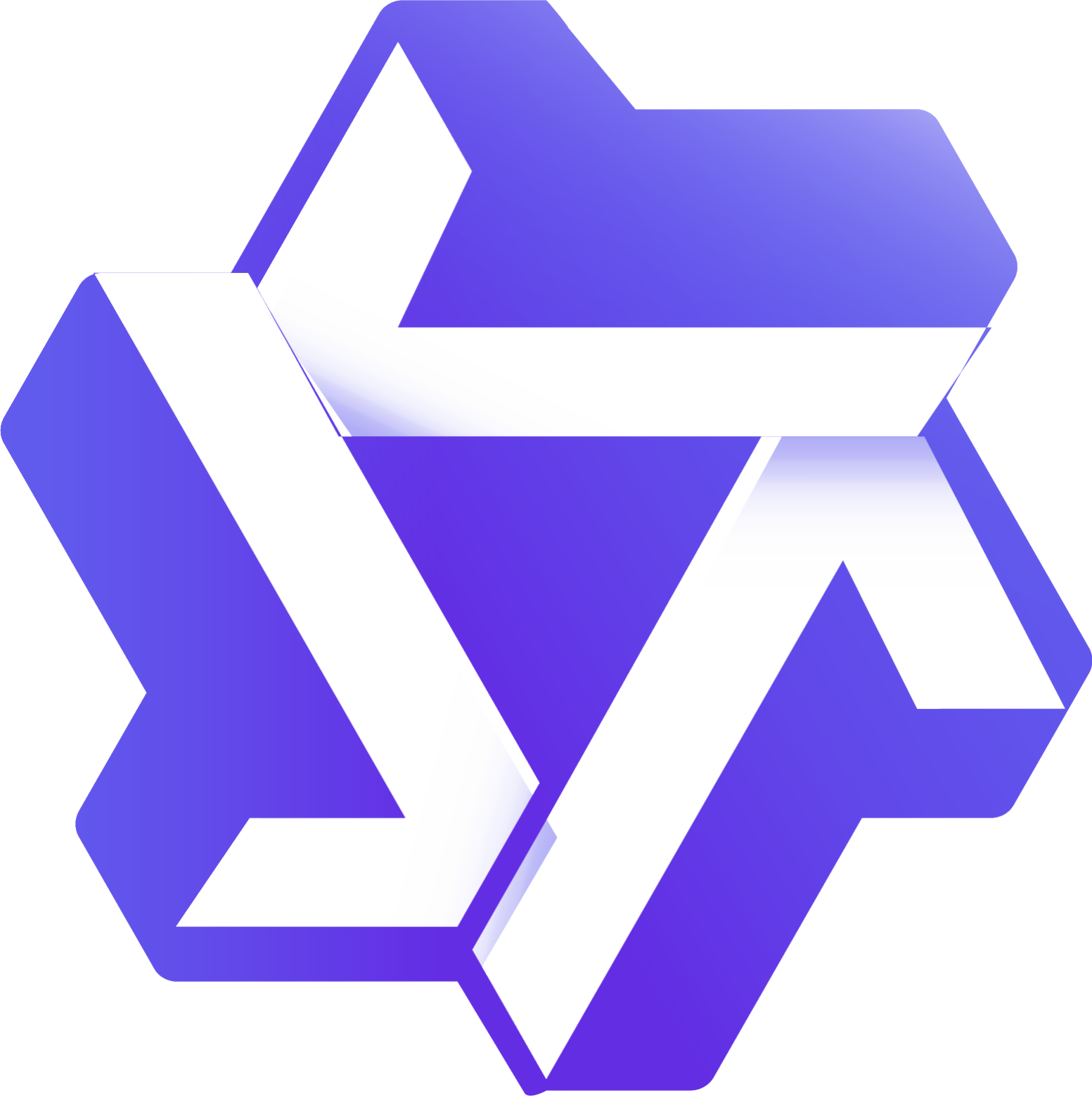}\,Qwen2.5-VL-7B & 31.4 & 15.2 & 34.4 & 16.5 & 35.6 & 16.7 & 31.6 & 15.5 & 26.2 & 13.1 \\
\midrule
\rowcolor{gray!15} \multicolumn{12}{l}{\textit{Training-free methods}} \\
VLFM$^\dagger$~{\scriptsize[ICRA'24]~\cite{vlfm}} & BLIP-2 & 12.8 & 7.5 & 16.9 & 9.7 & 15.3 & 8.9 & 11.6 & 6.7 & 9.9 & 6.1 \\
\multirow{2}{*}{3D-Mem~{\scriptsize[CVPR'25]~\cite{3dmem}}}
& \icon{image/qwen_logo.pdf}\,Qwen2.5-VL-3B & 15.3 & 2.8  & 20.7 & 3.7  & 20.3 & 4.0  & 13.4 & 2.3  & 10.2 & 1.8 \\
& \icon{image/qwen_logo.pdf}\,Qwen2.5-VL-7B & 21.2 & 8.7  & 21.3 & 8.4  & 18.8 & 8.1  & 21.7 & 8.8  & 22.7 & 9.2 \\
\midrule
\multirow{3}{*}{\textbf{Ours}}
& \icon{image/qwen_logo.pdf}\,Qwen2.5-VL-7B & 29.4 & 11.2 & 28.8 & 9.4  & 30.1 & 10.7 & 33.6 & 12.1 & 25.1 & 11.7 \\
& \icon{image/qwen_logo.pdf}\,Qwen3.5-9B    & 36.6 & 18.8 & 41.4 & 22.4 & 41.9 & 21.2 & 40.8 & 20.3 & 26.5 & 13.6
 \\
& \icon{image/qwen_logo.pdf}\,Qwen3-VL-235B-A22B-thinking & \textbf{39.6} & \textbf{19.9} & \textbf{43.7} & \textbf{22.5} & \textbf{47.7} & \textbf{23.7} & \textbf{43.8} & \textbf{21.4} & \textbf{26.8} & \textbf{14.0}
  \\
\bottomrule
\end{tabular}
}\caption{\textbf{Evaluation on LangMap using concise descriptions}.
Single-Goal reports the average performance over all single-goal episodes.
SAP-Nav achieves top-tier performance at all four granularity levels without task-specific training. VLFM$^\dagger$ is re-run with its official implementation.
To adapt VLFM to hierarchical OVON, BLIP-2 is prompted with granularity-specific goal information for frontier value estimation. The versions of GroundingDINO and SAM are kept consistent with those used by SAP-Nav.}
\label{tab:langmap}
    \vspace{-0.2in}
\end{table*}

\paragraph{Viewpoint Selection}
When the current observation is insufficient for reliable verification, we use viewpoint selection to identify a more informative viewpoint, combining geometric sampling, constraint filtering, and visibility scoring.
We first sample viewpoints $\mathcal{V}$ at uniformly spaced angles along concentric rings centered at $\mathbf{c}$ with radii $r_i\in\{0.8, 1.2, 1.6, 2.0, 2.4\}\,\mathrm{m}$ following the equation:
\begin{equation}
\mathbf{v}_{i,j}=\mathbf{c}+r_i\left(\cos\theta_j, \sin\theta_j\right),\qquad\theta_j=\frac{2\pi j}{N_{\theta}},
\end{equation}
where $j\in\{0,\ldots,N_{\theta}-1\}$, $N_{\theta}=24$ and $\mathbf{c}$ is the coordinatewise median of the segmentation footprint $\mathcal{U}(o)$ of the target on the obstacle map.
However, these geometric samples may not correspond to feasible robot
locations and visibility, we then filter them by two criteria:
(1) \emph{Navigable and explored:} located in navigable space within the explored region;
(2) \emph{Adaptive minimum distance:} to prevent low targets from falling outside the vertical field of view, the Euclidean distance from the viewpoint $\mathbf{v}$ to $\mathbf{c}$ should satisfy: 
\begin{equation}
\|\mathbf{v}-\mathbf{c}\|_2\geq\frac{h_{\mathrm{cam}} - h_{\mathrm{top}}}{\tan\left(\frac{\theta_v}{2}\right)}
\end{equation}
where $h_{\mathrm{cam}}$ is the camera height, $h_{\mathrm{top}}$ is the estimated top height of the candidate object, and $\theta_v$ denotes the vertical field of view of the camera.
Furthermore, we evaluate each $\mathbf{v}\in\mathcal{V}$ based on its visibility to the target.
To achieve this, we maintain an online 2.5D height map, where each explored BEV cell records the maximum observed obstacle height.
For each valid $\mathbf{v}\in\mathcal{V}$ and footprint cell $\mathbf{f}\in\mathcal{U}(o)$, we cast a ray from the camera center above $\mathbf{v}$ to the target point height at $\mathbf{f}$, which is deemed visible only if all intermediate cells of the ray lie below the resulting ray profile, denoted as $\mathrm{vis}(\mathbf{v},\mathbf{f})$.
The best viewpoint is determined as follows:
\begin{equation}
\mathbf{v}^{*}=\arg\max_{\mathbf{v}\in\mathcal{V}}\frac{1}{|\mathcal{U}(o)|}\sum_{\mathbf{f}\in\mathcal{U}(o)}\mathrm{vis}(\mathbf{v},\mathbf{f}).
\end{equation}
Upon reaching $\mathbf{v}^{*}$, the agent evaluates $s$ again. If $s$ remains below $\tau$ after three repositioning attempts, suggesting that the insufficiency may not primarily result from viewpoint quality, the agent will perform target verification using the observation with the highest $s$ collected so far.

\paragraph{Target Verification.}
If the sufficiency score reaches the threshold ($s \geq \tau$), the VLM is shown the highlighted candidate region in its surrounding context and asked whether it matches the target category. 
Furthermore, if $G$ contains attribute constraints, we first use an LLM to parse $G$ into a structured set of attribute descriptions~\cite{context}. For each category consistent candidate, the VLM then verifies whether its visual attributes match the parsed constraints and accepts the candidate only when all required attributes are satisfied.
Rejected candidates are added to a blacklist to avoid repeated VLM queries when they are detected again, after which the agent resumes exploration.

\section{Experiments}
\subsection{Experimental Setup}
\paragraph{Datasets.}
We evaluate on two OVON benchmarks built on realistic HM3D scans~\cite{hm3d} in the Habitat simulator~\cite{habitat}.
\textbf{LangMap}~\cite{langmap} is a large-scale, multi-granularity benchmark spanning 414 object categories across all HM3D-Sem~\cite{yadav2023habitat} validation scenes.
We adopt its single-goal protocol with around \textbf{15K} tasks at four linguistic granularity levels: \emph{scene-level} targets any instance of a category; \emph{room-level} constrains the target to a room type; \emph{region-level} further restricts it to a specific room instance; and \emph{instance-level} identifies a unique instance by its discriminative attributes.
\textbf{HM3D-OVON}~\cite{hm3dovon} is a popular scene-level object navigation benchmark with 379 open-vocabulary categories, where we evaluate on its Val Unseen split, which contains over \textbf{3K} episodes across 49 object categories.

\begin{table}[t!]
    \centering
    \renewcommand{\arraystretch}{1.25}
    \setlength{\tabcolsep}{8pt}
    \resizebox{\columnwidth}{!}{%
    \begin{tabular}{l@{\hspace{4pt}}c@{\hspace{5pt}}c@{\hspace{4pt}}}
    \toprule
    \textbf{Method} & \textbf{SR}\,$\uparrow$ & \textbf{SPL}\,$\uparrow$ \\
    \midrule
    \rowcolor{gray!15} \multicolumn{3}{l}{\textit{Training-based methods}} \\
    BC~{\scriptsize[IROS'24]~\cite{hm3dovon}}         & 5.4  & 1.9  \\
    DAgger~{\scriptsize[IROS'24]~\cite{hm3dovon}}     & 10.2 & 4.7  \\
    DAgRL~{\scriptsize[IROS'24]~\cite{hm3dovon}}      & 18.3 & 7.9  \\
    RL~{\scriptsize[IROS'24]~\cite{hm3dovon}}         & 18.6 & 7.5  \\
    DAgRL+OD~{\scriptsize[IROS'24]~\cite{hm3dovon}}   & 37.1 & 19.8 \\
    Uni-NaVid~{\scriptsize[RSS'25]~\cite{uninavid}}   & 39.5 & 19.8 \\
    MTU3D~{\scriptsize[ICCV'25]~\cite{mtu3d}}      & 40.8 & 12.1 \\
    Dynam3D~{\scriptsize[NeurIPS'25]~\cite{dynam3d}} & 42.7 & 22.4 \\
    OVSegDT~{\scriptsize[CVPR'26]~\cite{ovsegdt}}   & 44.7 & 20.6 \\
    NavFoM~{\scriptsize[ICLR'26]~\cite{navfom}}    & 45.2 & \textbf{31.9} \\
    \midrule
    \rowcolor{gray!15} \multicolumn{3}{l}{\textit{Training-free methods}} \\
    Modular GOAT~{\scriptsize[CVPR'24]~\cite{goat}} & 24.9 & 17.2 \\
    VLFM~{\scriptsize[ICRA'24]~\cite{vlfm}}       & 35.2 & 19.6 \\
    TANGO~{\scriptsize[CVPR'25]~\cite{tango}~(\icon{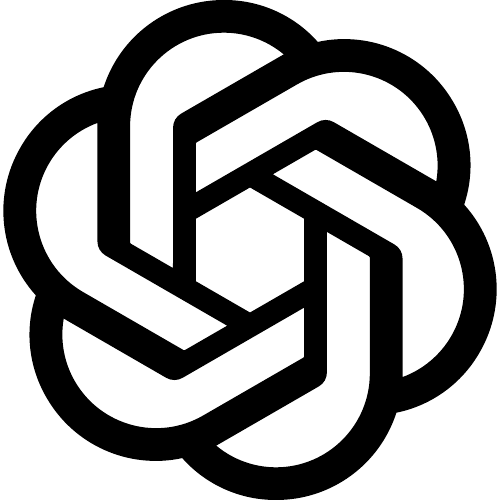}\,GPT-4o)}      & 35.5 & 19.5 \\
    OpenFrontier~{\scriptsize[arXiv'26]~\cite{openfrontier}~(Gemini-2.5-Flash)} & 39.0 & 20.1 \\
    MetaNav~{\scriptsize[arXiv'26]~\cite{metanav}~(\icon{image/ChatGPT_logo.pdf}\,GPT-4o)}   & 46.1 & 29.8 \\
    \midrule
    \textbf{Ours}~(\icon{image/ChatGPT_logo.pdf}\,GPT-4o) & \textbf{49.7} & 23.1 \\
    \bottomrule
    \end{tabular}%
    }
    \vspace{-0.1in}
    \caption{
Evaluation on the HM3D-OVON val unseen split.
Although designed for multi-granularity instructions, SAP-Nav achieves the top-ranked SR among all compared methods on standard scene-level OVON without navigation training.}
    \label{tab:hm3d-ovon}
    \vspace{-0.2in}
    \end{table}

\paragraph{Metrics.}
    Following~\cite{habitatchallenge2023}, we report two standard metrics: 
    (1) Success Rate (SR), which measures the fraction of episodes that meet the success criteria, $\mathrm{SR} = \frac{1}{N}\sum_{i=1}^{N} S_i$; and
    (2) Success weighted by Path Length (SPL), which further rewards path efficiency, $\mathrm{SPL} = \frac{1}{N}\sum_{i=1}^{N} S_i \frac{l_i}{\max(p_i, l_i)}$.
    Following both benchmarks, an episode succeeds if the agent reaches any dataset-annotated target viewpoint within 1\,m at a maximum of 500 steps. 

\paragraph{Implementation Details.}
The action space $A$ includes \textls[50]{\footnotesize\textsc{MOVE\_FORWARD}} (0.25\,m), \textls[50]{\footnotesize\textsc{TURN\_LEFT}}/\textls[50]{\footnotesize\textsc{TURN\_RIGHT}} (30$^{\circ}$) and \textls[50]{\footnotesize\textsc{STOP}}. 
For \emph{VLM-based reasoning}, we use
(1) open-source VLMs, Qwen2.5-VL-7B-Instruct~\cite{qwen25vl}, Qwen3.5-9B enabled thinking mode~\cite{qwen35}, Qwen3-VL-235B-A22B-thinking~\cite{Qwen3-VL}; and 
(2) the closed-source VLM GPT-4o~\cite{gpt4o}.
\subsection{Benchmark Results}
\paragraph{Hierarchical OVON task on LangMap.}
Table 1 reports the outstanding performance of SAP-Nav in tackling hierarchical OVON.
SAP-Nav with Qwen3-VL-235B-A22B-thinking achieves the SOTA performance at every granularity level without task-specific navigation training. With Qwen3.5-9B, SAP-Nav already exceeds the strongest training-based baseline, PlaNaVid. Notably, even with the same Qwen2.5-VL-7B backbone, SAP-Nav also surpasses PlaNaVid on the Region level.
Compared with training-free methods, SAP-Nav substantially outperforms the adapted VLFM baseline. 
Under the same Qwen2.5-VL-7B backbone as 3D-Mem, SAP-Nav improves performance across all granularity levels, with the largest gains on room- and region-levels. 

\paragraph{OVON task on HM3D-OVON.}
We further evaluate SAP-Nav on the widely used HM3D-OVON benchmark to examine its generalization to standard scene-level OVON.
As shown in Table \ref{tab:hm3d-ovon}, SAP-Nav achieves the highest SR among all compared methods.
This indicates that SAP-Nav, although designed for multi-granularity instructions, transfers well to conventional object-goal navigation.
In terms of SPL, SAP-Nav is lower than NavFoM and MetaNav.
We attribute this to the AVV module, which deliberately repositions the agent to verify candidate targets when the current observation is insufficient; the additional movements lengthen the traversed path but substantially improve reliability.
We consider this a reasonable trade-off, as successfully reaching the correct target is typically more critical than path efficiency in real-world deployment.

\subsection{Real-world Deployment}
\begin{figure*}[htpb]
        \centering
        \begin{adjustbox}{center}
            \includegraphics[width=\textwidth]{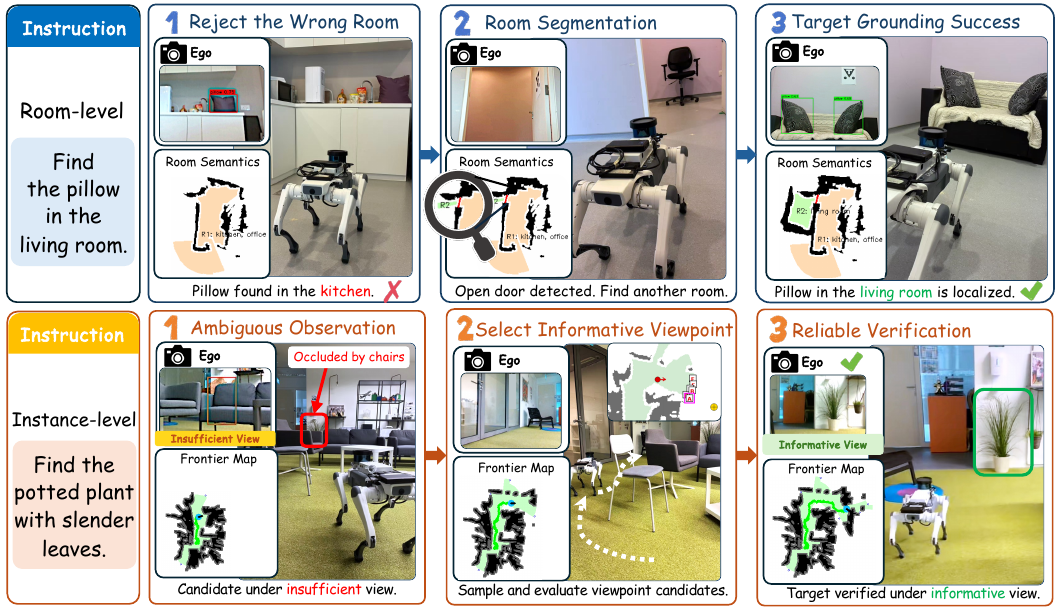}
        \end{adjustbox}
        \vspace{-0.2in}
        \caption{\textbf{Real-world deployment}. SAP-Nav can execute hierarchical OVON instructions in indoor scenes.}
        \label{fig:real-world}
    \end{figure*}

We validate SAP-Nav on a Deep Robotics Jueying Lite3 quadruped equipped with an RGB-D camera and a LiDAR sensor. 
Onboard sensing and navigation processes communicate with a remote RTX A5000 workstation over WiFi. SAP-Nav runs on the workstation for online inference, and SLAM Toolbox provides localization while Nav2 handles point-to-point navigation with dynamic obstacle avoidance. 
We conduct real-robot experiments across representative indoor environments, including offices, meeting rooms, living rooms, and kitchens, using diverse instructions spanning multiple semantic granularities and navigation goals.
Fig.~\ref{fig:real-world} shows qualitative examples of SAP-Nav in real indoor environments.
The upper row shows QSSR rejecting spatially inconsistent candidates through online room semantics, while the lower row shows AVV improving the viewpoint before target verification.
These examples illustrate the ability of SAP-Nav to support hierarchical OVON on a physical robot.

\begin{table}[t]
    \centering
    \renewcommand{\arraystretch}{0.75}
    \scriptsize
    \resizebox{\columnwidth}{!}{%
    \begin{tabular}{ccc cc cc}
        \toprule
        \multicolumn{3}{c}{\textbf{Component}}
        & \multicolumn{2}{c}{\textbf{Room}}
        & \multicolumn{2}{c}{\textbf{Region}} \\
        \cmidrule(lr){1-3}
        \cmidrule(lr){4-5}
        \cmidrule(lr){6-7}

        $S_1$ & $S_2$ & $S_3$
        & SR\,$\uparrow$ & SPL\,$\uparrow$
        & SR\,$\uparrow$ & SPL\,$\uparrow$ \\
        \midrule

        & & 
        & 26.9 & 14.3
        & 32.3 & 17.0 \\

        & & \checkmark
        & 39.4 & 20.2
        & 37.2 & 16.6 \\

        \checkmark & \checkmark &
        & 45.8 & 23.2
        & 42.3 & 19.8 \\

        \checkmark & \checkmark & \checkmark
        & \textbf{47.7} & \textbf{23.7}
        & \textbf{43.8} & \textbf{21.4} \\

        \bottomrule
    \end{tabular}%
    }
    \caption{Ablation study of QSSR across Room and Region levels on LangMap.}
    \label{tab:ablation1}
    \vspace{-0.2in}
\end{table}

\begin{table*}[t!]   
 \centering
    \renewcommand{\arraystretch}{0.9}
    \tiny
    \resizebox{\textwidth}{!}{%
    \begin{tabular}{ccc cc|cc cc cc cc | cc}
        \toprule
        \multicolumn{3}{c}{\textbf{Component}}
        & \multicolumn{10}{c|}{\textbf{LangMap}}
        & \multicolumn{2}{c}{\textbf{HM3D-OVON}} \\
        \cmidrule(r){1-3}
        \cmidrule(lr){4-13}
        \cmidrule(l){14-15}

        \textbf{SE} & \textbf{VS} & \textbf{TV}
        & \multicolumn{2}{c|}{\textbf{Single-Goal}}
        & \multicolumn{2}{c}{\textbf{Scene}}
        & \multicolumn{2}{c}{\textbf{Room}}
        & \multicolumn{2}{c}{\textbf{Region}}
        & \multicolumn{2}{c|}{\textbf{Instance}}
        & \multicolumn{2}{c}{\textbf{Overall}} \\
        \cmidrule(lr){4-5}
        \cmidrule(lr){6-7}
        \cmidrule(lr){8-9}
        \cmidrule(lr){10-11}
        \cmidrule(lr){12-13}
        \cmidrule(lr){14-15}

        & & 
        & SR\,$\uparrow$ & SPL\,$\uparrow$
        & SR & SPL
        & SR & SPL
        & SR & SPL
        & SR & SPL
        & SR & SPL \\
        \midrule

        & & 
        & 24.1 & 12.2
        & 12.8 & 7.2
        & 38.3 & 20.7
        & 40.0 & 18.9
        & 3.5 & 1.7
        & 24.5 & 16.2 \\

        \checkmark & \checkmark &
        & 29.8 & 15.7
        & 17.6 & 10.9
        & 39.8 & 20.7
        & 40.5 & 20.4
        & 17.7& 9.6
        & 31.1 & 17.5 \\

        & & \checkmark
        &  36.0& 19.2
        & 33.1 & 19.9
        & 44.0 & 22.7
        &  42.5& \textbf{22.0}
        & 25.0 & 13.4
        & 46.8 & \textbf{26.2} \\

        & \checkmark & \checkmark
        & 39.2 & 17.6
        & 43.5 & 20.3
        &46.8 & 21.4
        & \textbf{44.0} & 17.6
        & 26.3 & 13.1
        & 49.4 &22.0 \\

        \checkmark & \checkmark & \checkmark
        & \textbf{39.6} & \textbf{19.9} & \textbf{43.7} & \textbf{22.5} & \textbf{47.7} & \textbf{23.7} & 43.8 & 21.4 & \textbf{26.8} & \textbf{14.0}
        & \textbf{49.7} & 23.1 \\

        \bottomrule
    \end{tabular}%
    }
    \vspace{-0.1in}
    \caption{Ablation study on the three components of AVV across all granularities of LangMap and HM3D-OVON benchmarks.}
    \label{tab:ablation2}
    \vspace{-0.2in}
\end{table*}

\subsection{Ablation Study}

\paragraph{Effect of Queryable Spatial Semantic Representation}
As shown in Table~\ref{tab:ablation1}, we ablate QSSR to isolate the roles of holistic room semantics and sub-room semantic refinement.
$S_1$ is not evaluated alone, as it serves as the structural basis for $S_2$ and only provides geometric segmentation without semantics.
Removing QSSR entirely causes the most severe degradation, since the agent cannot handle room- or region-level constraints in the instruction without maintaining spatial semantic awareness of the explored environment.
Using only $S_3$, the agent relies on scene predictions accumulated from passive observations. 
The lower performance indicates that individual views provide limited context and can inject noisy room-type evidence into $\Phi(\mathbf{u})$.
The $S_1+S_2$ variant removes $S_3$ and thus lacks sub-room refinement, which hurts performance in open areas or under-segmented rooms where geometric room boundaries do not match functional areas. 

\paragraph{Effect of Active Viewpoint Verification} 
Table~\ref{tab:ablation2} shows that AVV improves performance through sufficiency evaluation (SE), viewpoint selection (VS), and target verification (TV).
Removing AVV causes a clear performance drop, indicating that passive observations are insufficient for reliable stopping.
SE and VS improve performance by acquiring more informative observations, while TV brings larger gains by explicitly checking the candidate against the instruction. 
Comparing TV with VS+TV shows that viewpoint quality is important for verification, although the extra repositioning caused by VS can reduce the path efficiency, as reflected by the lower SPL.
SE mitigates this loss by judging whether the current observation is already sufficient, so that VS is triggered only when needed.

\section{Conclusion}
We introduced SAP-Nav, a fully online, zero-shot navigation framework for hierarchical OVON. 
Our key design insight is that spatial semantic understanding and target verification in this task both require distinct visual evidence that passive observation cannot guarantee.
To address this challenge, QSSR constructs a queryable spatial semantic representation by extracting explicit room type semantics from actively acquired holistic room snapshots, while AVV assesses the current visual evidence and seeks a more informative viewpoint when necessary before verifying the target category and attributes.
Extensive experiments across two benchmarks and real-world robotic deployments demonstrated that SAP-Nav effectively handled multi-granularity instructions and demonstrated its strong practical feasibility in unseen environments.

\vspace{0.05in}
\noindent\textbf{Limitations and Future Work.}
While active perception improves reliability in hierarchical OVON, SAP-Nav can be further improved in two directions.
First, although 
the designed sufficiency evaluation avoids unnecessary repositioning, AVV selects viewpoints based on geometric visibility, without considering their semantic utility or motion cost. Future work could jointly optimize semantic informativeness, verification reliability, and navigation efficiency.
Second, QSSR is not retained across successive tasks. Integrating it with persistent memory could enable acquired spatial semantic knowledge to be reused for more efficient lifelong navigation.

\clearpage
\bibliography{aaai2027}

\end{document}